\documentclass[twoside,11pt]{article}

\usepackage{jmlr2e}
\usepackage{pythonhighlight}
\usepackage{times}
\usepackage{hyperref}
\usepackage{helvet}
\usepackage{courier}
\usepackage{natbib}
\usepackage{verbatim}
\usepackage{makeidx}  
\usepackage{dsfont, times}
\usepackage{algorithm}
\usepackage{algorithmic}
\usepackage{amsfonts}
\usepackage{amssymb}
\usepackage{amsmath}
\usepackage{graphicx}
\usepackage{booktabs}

\usepackage{color}
\usepackage{url}
\usepackage{listings}

\usepackage{bm}
\usepackage[american]{babel}

\DeclareMathOperator*{\argmin}{argmin}

\newcommand{\picspace}{0.245}

\newcommand{\expwidth}{1.6}

\jmlrheading{1}{2021}{1-48}{4/00}{10/00}{Liu, Hu, Qian, Yu and Qian}

\ShortHeadings{ZOOpt: Toolbox for Derivative-Free Optimization}{Liu, Hu, Qian, Yu and Qian}
\firstpageno{1}
\begin{document}

\title{ZOOpt: Toolbox for Derivative-Free Optimization}

\author{\name Yu-Ren Liu \email liuyr@lamda.nju.edu.cn \\
       \name Yi-Qi Hu \email huyq@lamda.nju.edu.cn \\
       \name Chao Qian \email qianc@nju.edu.cn \\
       \name Yang Yu\thanks{Correspondence author} \email yuy@nju.edu.cn \\
       \addr National Key Laboratory for Novel Software Technology\\
       Nanjing University, Nanjing 210023, China
       \AND
       \name Hong Qian \email hqian@cs.ecnu.edu.cn \\
       \addr School of Computer Science and Technology \\East China Normal University, Shanghai  200062, China}

\editor{}

\maketitle

\begin{abstract}
Recent advances in derivative-free optimization allow efficient approximation of the global-optimal solutions of sophisticated functions, such as functions with many local optima, non-differentiable and non-continuous functions.  This article describes the ZOOpt (\textbf{Z}eroth \textbf{O}rder \textbf{Opt}imization) toolbox that provides efficient derivative-free solvers and is designed easy to use. ZOOpt provides single-machine parallel optimization on the basis of python core and multi-machine distributed optimization for time-consuming tasks by incorporating with the Ray framework --- a famous platform for building distributed applications. ZOOpt particularly focuses on optimization problems in machine learning, addressing high-dimensional and noisy problems such as hyper-parameter tuning and direct policy search. The toolbox is maintained toward a ready-to-use tool in real-world machine learning tasks.
\end{abstract}


\section{Derivative-Free Optimization}
Optimization, taking $x^* = \argmin_{\bm{x} \in \mathcal{X}}f(\bm{x})$ as a general representative, is fundamental in machine learning. Derivative-free optimization, also termed as zeroth-order or black-box optimization, involves a class of optimization algorithms that do not rely on gradient information. 
In recent years, derivative-free optimization has achieved remarkable applications in machine learning, including hyper-parameter optimization~\citep{KDD13autoweka,NIPS15autosklearn}, direct policy search~\citep{SalimansHCS17,hu2017sequential}, subset selection~\citep{qian.yu.nips15}, image classification~\citep{ICicml17}, etc.
Representative derivative-free algorithms include evolutionary
algorithms~\citep{HansenMK03}, Bayesian optimization~\citep{reviewBO16}, optimistic optimization~\citep{MunosFTML2014}, model-based optimization~\citep{yu.qian.racos}, etc.

\section{Classification-based Optimization}
Model-based derivative-free optimization algorithms share a framework that iteratively learns a model for promising search areas and samples solutions from the model. Different kinds of methods usually vary in the design of the model. For example, cross-entropy methods \citep{Boer2005} may use Gaussian distribution as the model, Bayesian optimization methods \citep{SnoekLA2012} employ Gaussian process to model the joint distribution, and the estimation of distribution algorithms have incorporated many kinds of learning models. Classification-based optimization algorithms learn a particular type of model: classification model, leading to theoretical grounded properties of optimization performance. A classification model learns to classify solutions into two categories, $good$ or $bad$. Then solutions are sampled from the $good$ areas. \textsc{SRacos} \citep{hu2017sequential}  is a recently proposed classification-based optimization algorithm. Unlike other model-based optimization algorithms, the sampling region of \textsc{SRacos} is learned by a simple classifier, which maintains an axis-parallel rectangle to cover all the positive but no negative solutions. \textsc{SRacos} shows outstanding performance in empirical studies. With the aim of supporting machine learning tasks, ZOOpt implements a set of classification-based methods  that are efficient and performance-guaranteed, with add-ons handling noise and high-dimensionality.

\section{Methods in ZOOpt}

\begin{table}[bth]
    \centering
    \includegraphics[width=\textwidth]{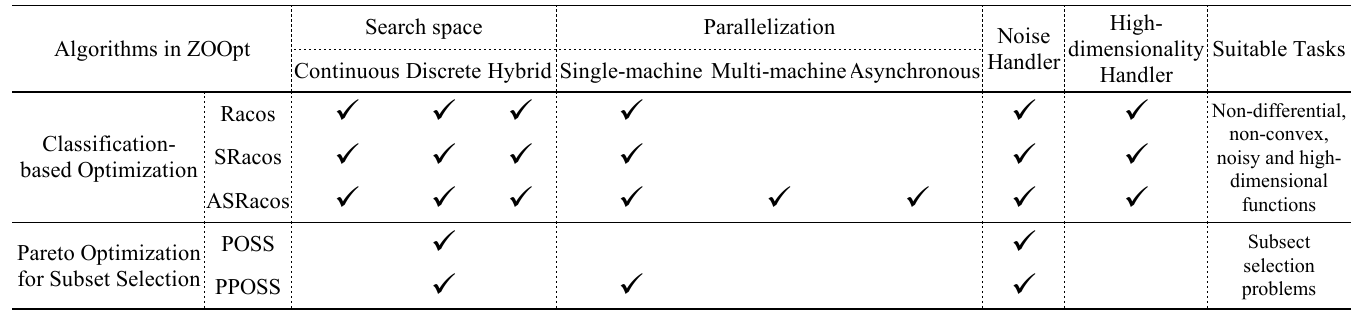}
	\caption{Algorithms implemented in the ZOOpt toolbox. For each algorithm, we conclude its support on different kinds of search space, parallelization and the compatibility with the noise handler and the high-dimensional handler.}
	\label{sfunctions}
\end{table}
{\bf Optimization in the continuous/discrete/hybrid space}. 
We implement \textsc{SRacos}~\citep{hu2017sequential} as the default optimization method, which has shown high efficiency in a range of learning tasks. Optional methods are \textsc{Racos}~\citep{yu.qian.racos} and \textsc{ASRacos}~\citep{Asracos2019}, respectively are the batch and asynchronous version of \textsc{SRacos}. A routine is in place to setup the default parameters of the two methods, while users can override them. Benefit from the compatibility of the classifier with multiple data types, classification-based optimization supports optimization in the continuous, discrete (categorical), or hybrid space naturally. 

{\bf Optimization in the binary vector space with constraint}.
If the optimization task is in a binary vector space with constraints, such as the subset selection problem, POSS~\citep{qian.yu.nips15} is the default optimization method. POSS treats subset selection task as a bi-objective optimization problem that simultaneously optimizes some given criterion and the subset size. POSS has been proven with the best-so-far approximation quality on these problems. PPOSS ~\citep{DBLP:conf/ijcai/QianSYTZ16} is the parallel version of the POSS algorithm. 

{\bf Noise handling}. 
Noise has a great impact on the performance of derivative-free optimization. Resampling is the most straightforward method to handle noise, which evaluates one sample several times to obtain a stable mean value. Besides resampling, more efficient methods including value suppression~\citep{AAAI18noise} and  threshold selection~\citep{NIPS17noise} are implemented.

{\bf High-dimensionality handling}.
Increase of the search space dimensionality badly injures the performance of derivative-free optimization. When a high dimensional search space has a low effective-dimension, random embedding~\citep{Wang2016remboJAIR} is an effective way to improve the efficiency. Also, the sequential random embeddings~\citep{IJCAI16sre} can be used when there is no clear low effective-dimension.

{\bf Distributed Optimization}.
Evaluation of a sampled solution is usually time consuming for many real-world optimization tasks, such as hyperparameter tuning in large-scale machine learning projects. Incorparating with the Ray framework ~\citep{MoritzNWTLLEYPJ18}, ZOOpt implements an efficient distributed optimization module that enables users to parallelize single-machine code, with little to zero code changes.

\section{Usage}
In this section, we will briefly introduce the single-machine optimization, distributed optimization, optimization under noise and optimization in the high dimensional space through a few examples. For the full tutorial, including the detailed API introduction, hyper-parameter tuning tricks and all examples, we refer readers to \url{https://zoopt.readthedocs.io/en/latest/}.

{\bf Single-machine optimization}.
The core architecture of ZOOpt includes three parts: $Objective$, $Parameter$ and $Opt.min$. The $Objective$ object defines the function expression and the search space. The $Parameter$ object defines all parameters used by the optimization algorithm. $Opt.min$ is the interface for performing optimization. After defining a user-specified objective function and the corresponding search space, only one line of code is needed to perform optimization by using $Opt.min$.  A quick-start example is provided as follows. 

\begin{python}
import numpy as np
from zoopt import ValueType, Dimension2, Objective, Parameter, Opt

def ackley(solution):
        x = solution.get_x()
        bias = 0.2
        value = -20 * np.exp(-0.2 * np.sqrt(sum([(i - bias) * (i - bias) for i in x]) / len(x))) - \
                np.exp(sum([np.cos(2.0*np.pi*(i-bias)) for i in x]) / len(x)) + 20.0 + np.e
        return value

dim_size = 100  # dimension size
dim = Dimension2([(ValueType.CONTINUOUS, [-1, 1], 1e-6)]*dim_size)
obj = Objective(ackley, dim)
# perform optimization
solution = Opt.min(obj, Parameter(budget=100*dim_size))
# print the solution
print(solution.get_x(), solution.get_value())
# parallel optimization for time-consuming tasks
solution = Opt.min(obj, Parameter(budget=100*dim_size, parallel=True, server_num=3))
\end{python}

{\bf Distributed optimization}.
Distributed optimization in ZOOpt is implementd by incorparating with Ray. Currently, ZOOpt is an optional optimization tool  in $Ray.tune$ --  a library for fast hyperparameter tuning at any scale. Through $Ray.tune$, users can easily distribute the optimization without caring about the communication infrastructure. We provide an example as follows.
\begin{python}
import time
from ray import tune
from ray.tune.suggest.zoopt import ZOOptSearch
from ray.tune.schedulers import AsyncHyperBandScheduler
from zoopt import ValueType  # noqa: F401

def evaluation_fn(step, width, height):
    time.sleep(0.1)
    return (0.1 + width * step / 100)**(-1) + height * 0.1

def easy_objective(config):
    # Hyperparameters
    width, height = config["width"], config["height"]

    for step in range(config["steps"]):
        # Iterative training function - can be any arbitrary training procedure
        intermediate_score = evaluation_fn(step, width, height)
        # Feed the score back back to Tune.
        tune.report(iterations=step, mean_loss=intermediate_score)

if __name__ == "__main__":
    import argparse
    parser = argparse.ArgumentParser()
    parser.add_argument(
        "--smoke-test", action="store_true", help="Finish quickly for testing")
    parser.add_argument(
        "--server-address",
        type=str,
        default=None,
        required=False,
        help="The address of server to connect to if using "
        "Ray Client.")
    args, _ = parser.parse_known_args()

    if args.server_address:
        import ray
        ray.init(f"ray://{args.server_address}")
    num_samples = 10 if args.smoke_test else 1000
    # Optional: Pass the parameter space yourself
    # space = {
    #     # for continuous dimensions: (continuous, search_range, precision)
    #     "height": (ValueType.CONTINUOUS, [-10, 10], 1e-2),
    #     # for discrete dimensions: (discrete, search_range, has_order)
    #     "width": (ValueType.DISCRETE, [0, 10], True)
    #     # for grid dimensions: (grid, grid_list)
    #     "layers": (ValueType.GRID, [4, 8, 16])
    # }
    zoopt_search_config = {
        "parallel_num": 8,
    }
    zoopt_search = ZOOptSearch(
        algo="Asracos",  # only support ASRacos currently
        budget=num_samples,
        # dim_dict=space,  # If you want to set the space yourself
        **zoopt_search_config)
    scheduler = AsyncHyperBandScheduler()
    analysis = tune.run(
        easy_objective,
        metric="mean_loss",
        mode="min",
        search_alg=zoopt_search,
        name="zoopt_search",
        scheduler=scheduler,
        num_samples=num_samples,
        config={
            "steps": 10,
            "height": tune.quniform(-10, 10, 1e-2),
            "width": tune.randint(0, 10)
        })
    print("Best config found: ", analysis.best_config)
\end{python}

{\bf Optimization under noise}. The noise handler can be enabled through adding some attributes to the definition of the $Parameter$ object. Three kinds of noise handlers are implemented in ZOOpt. Naive re-sampling reduces the noise by evaluating the same solution for many times and taking their mean value as the final result. Value suppression ~\citep{AAAI18noise} reduces the noise with a higher efficiency by re-evaluating the best solution when it isn't updated for a pre-defined number of times. Threshold selection ~\citep{NIPS17noise} is a noise handler customized for the POSS algorithm, where the solution $x$ is better than $y$ only if $f(x)$ is smaller than $f(y)$ by at least a threshold. We provide simplified cases on how to use these noise handlers as follows. Their full versions can be found in the tutorial.

\begin{python}
from zoopt import Parameter
from sparse_mse import SparseMSE
import numpy as np

# naive resampling
parameter = Parameter(budget=200000, noise_handling=True, resampling=True, resample_times=10)
# value suppression
parameter = Parameter(budget=200000, noise_handling=True, suppression=True, non_update_allowed=500, resample_times=100, balance_rate=0.5)
# threshold selection
mse = SparseMSE('sonar.arff')
mse.set_sparsity(8)
parameter = Parameter(algorithm='poss', noise_handling=True, ponss=True, ponss_theta=0.5, ponss_b=mse.get_k(), budget=2 * np.exp(1) * (mse.get_sparsity() ** 2) * mse.get_dim().get_size())
\end{python}

{\bf Optimization in the high-dimensional space.}
ZOOpt contains a high-dimensionality handling algorithm called sequential random embedding (SRE) ~\citep{IJCAI16sre}. SRE runs the optimization algorithms in the low-dimensional space, where the function values of solutions are evaluated via the embedding into the original high-dimensional space sequentially. SRE is effective for the function class that all dimensions may affect the function value but many of them only have a small bounded effect, and can scale \textsc{Racos}, \textsc{SRacos} and \textsc{ASRacos} (the main optimization algorithm in ZOOpt) to 100,000-dimensional problems. The high-dimensionality handler can be enabled through adding  attributes to the definition of the $Parameter$ object. An example is provided as follows.

\begin{python}
from simple_function import sphere_sre
from zoopt import Dimension, ValueType, Dimension2, Objective, Parameter, ExpOpt
dim_size = 10000  # dimension size
dim_regs = [[-1, 1]] * dim_size  # search space
dim_tys = [True] * dim_size  # continuous
dim = Dimension(dim_size, dim_regs, dim_tys)  # form up the dimension object
objective = Objective(sphere_sre, dim)  # form up the objective function
budget = 2000  # number of calls to the objective function
parameter = Parameter(budget=budget, high_dimensionality_handling=True, reducedim=True, num_sre=5, low_dimension=Dimension(10, [[-1, 1]] * 10, [True] * 10))
solution_list = ExpOpt.min(objective, parameter, repeat=1, plot=True)
\end{python}

\section{Experiments}
In our experiments, we aim to answer following questions:
(1) How does ZOOpt compare to prior derivative-free optimization toolboxes in classic optimization benchmarks? (2) Can ZOOpt scale better than other toolboxes when the dimension size of the optimization task increases? (3) Can ZOOpt have better robustness against noise than other toolboxes? (4) How does ZOOpt compare to other toolboxes in machine learning tasks?

To answer those questions, we compare ZOOpt to several prior derivative-free optimization toolboxes, incuding pycma \footnote{\url{https://github.com/CMA-ES/pycma}}, DEAP \footnote{\url{https://github.com/DEAP/deap}}, pygad \footnote{\url{https://github.com/ahmedfgad/GeneticAlgorithmPython}} and Hyperopt \footnote{\url{https://github.com/hyperopt/hyperopt}}. Pycma ~\citep{hansen2019pycma} is a Python implementation of the CMA-ES ~\citep{HansenMK03} algorithm. DEAP  ~\citep{DEAP_JMLR2012} is a evolutionary computation framework. Pygad ~\citep{gad2021pygad} is an open-source Python library of  genetic algorithms. Hyperopt ~\citep{DBLP:conf/icml/BergstraYC13} implements the state-of-the-art Bayesian optimization algorithms for hyperparameter tuning. For all toolboxes, we choose the default algorithm and the recommended parameters according to their tutorials.  It’s worth noting that each toolbox actually implements many optimization algorithms. However we don't exhaust the algorithm-level comparison in this paper, instead, we choose the default algorithm and focus more on the toolbox itself.  We refer readers who are interested in the algorithm-level comparision to the paper \textsc{Racos}~\citep{yu.qian.racos}, \textsc{SRacos}~\citep{hu2017sequential}, \textsc{ASRacos}~\citep{Asracos2019}, POSS~\citep{qian.yu.nips15} and PPOSS~\citep{DBLP:conf/ijcai/QianSYTZ16}.  Source code of the experiments can be found from \url{https://github.com/AlexLiuyuren/ZOOpt\_experiment}.

Experiments are conducted on three kinds of tasks. To answer question (1), (2), (3), we conduct experiments on optimizing benchmark synthetic functions. We empirically evaluate the performance of the tested toolboxes, including the convergence rate, the scalability and the robustness against noise, on four benchmark synthetic functions. To answer question (4), we then conduct experiments on two machine learning tasks. We study on a classification task with Ramploss, where the objective function is similar to that of support vector machines (SVM) but the loss  funciton of SVM is the hinge loss. We then study on the direct policy search for controlling tasks, where the policy is a fully connected feedforward neural network and its weights are optimized directly by derivative-free optimization algorithms.

\begin{figure}[h]
	\begin{minipage}[t]{\picspace\linewidth}
		\centering
		\includegraphics[width=\textwidth ]{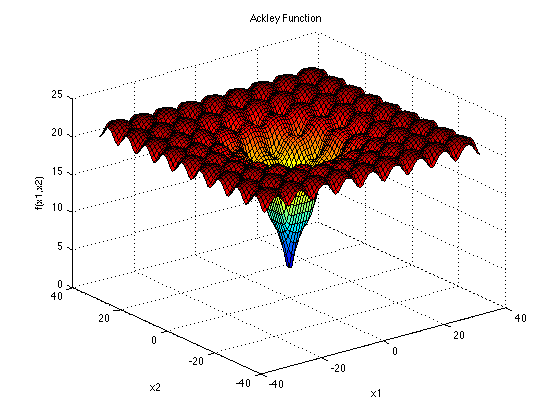}
		(a) Ackley
	\end{minipage}
	\begin{minipage}[t]{\picspace\linewidth}
		\centering
		\includegraphics[width=\textwidth]{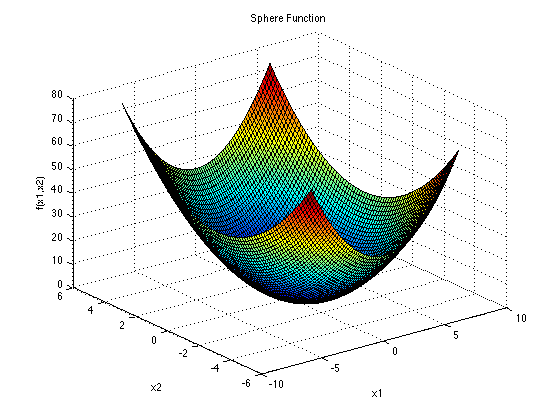}
		(b) Sphere
	\end{minipage}
	\begin{minipage}[t]{\picspace\linewidth}
		\centering
		\includegraphics[width=\textwidth]{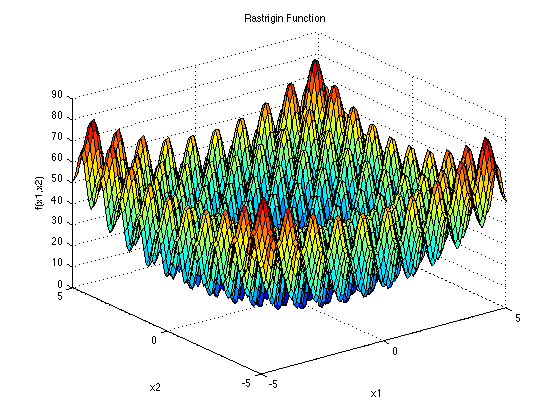}
		(c) Rastrigin
	\end{minipage}
	\begin{minipage}[t]{\picspace \linewidth}
		\centering
		\includegraphics[width=\textwidth]{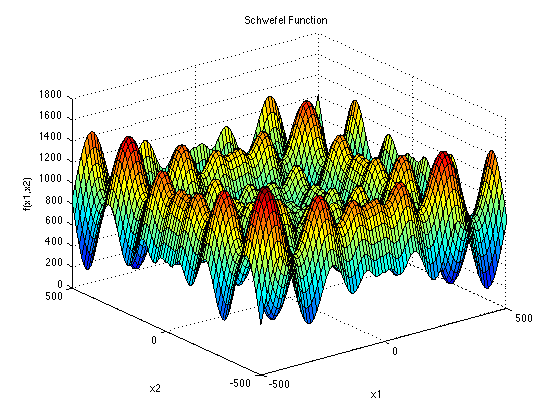}
		(d) Schwefel
	\end{minipage}
	\caption{3-d graphs of four Benchmark synthetic functions ( from \url{http://www.sfu.ca/\%7Essurjano/optimization.html}).  Among them, the Ackley, Rastrigin and Schwefel functions are highly non-convex while the Sphere function is conve. }
	\label{sfunctions}
\end{figure}
\subsection{Results on optimizing synthetic functions}
\begin{figure*}[t]
	\begin{minipage}[t]{\picspace \linewidth}
		\centering
		\includegraphics[width=\expwidth in]{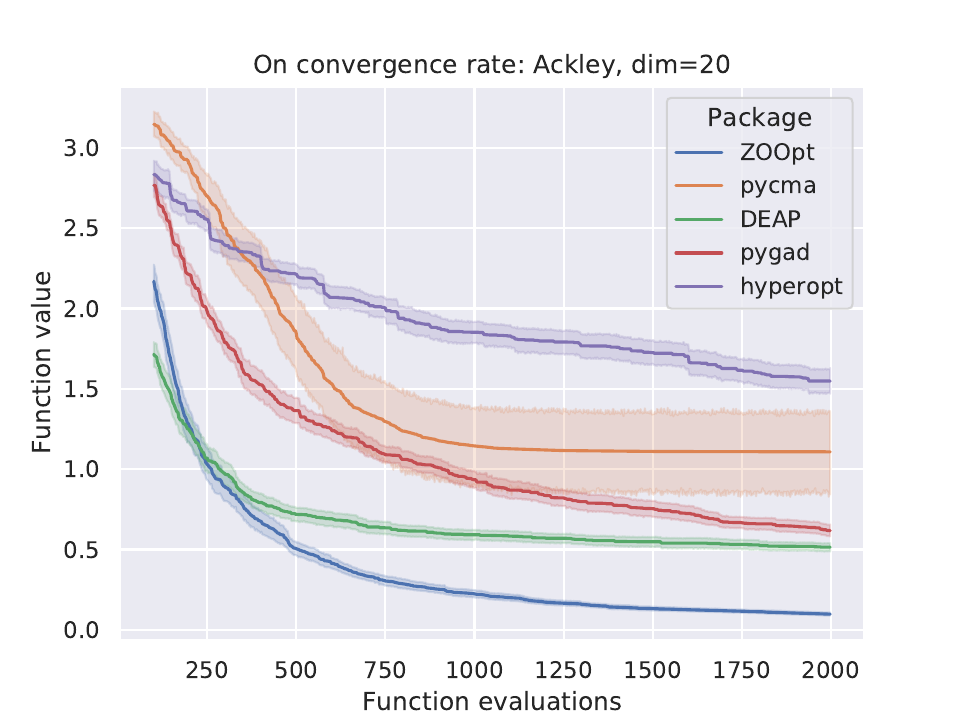}
	\end{minipage}%
	\begin{minipage}[t]{\picspace \linewidth}
		\centering
		\includegraphics[width=\expwidth in]{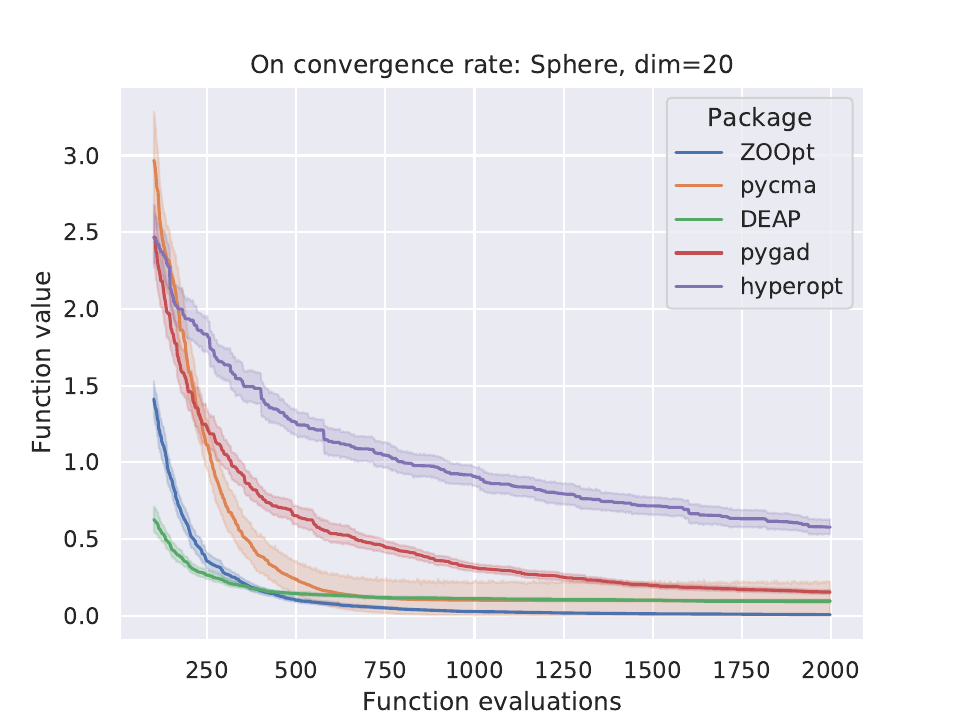}
	\end{minipage}
	\begin{minipage}[t]{\picspace\linewidth}
		\centering
		\includegraphics[width=\expwidth in]{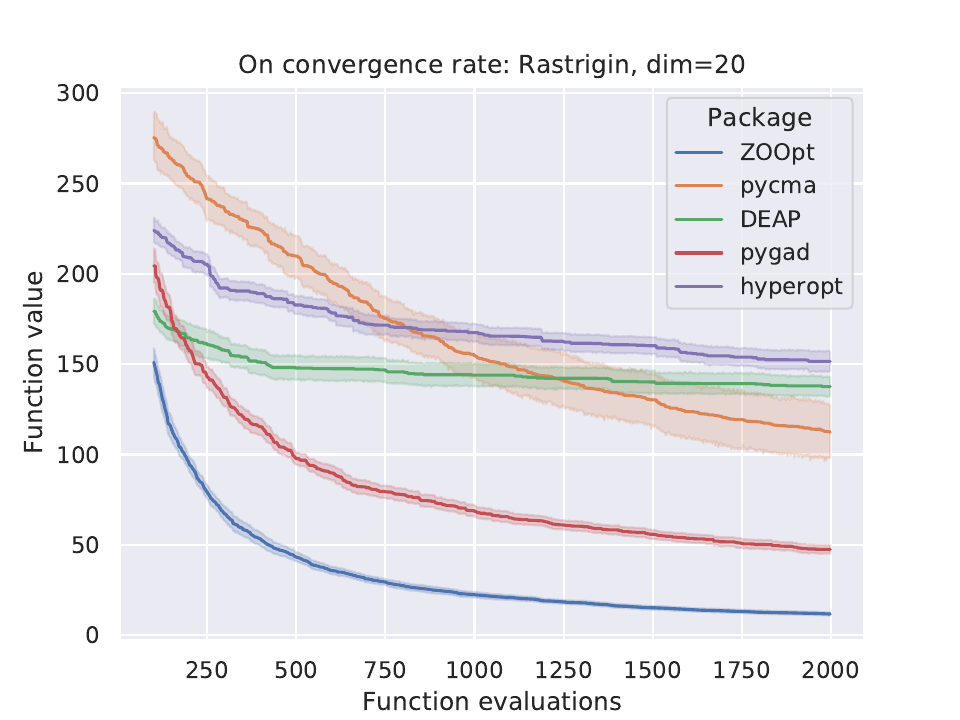}
	\end{minipage}
	\begin{minipage}[t]{\picspace \linewidth}
		\centering
		\includegraphics[width=\expwidth in]{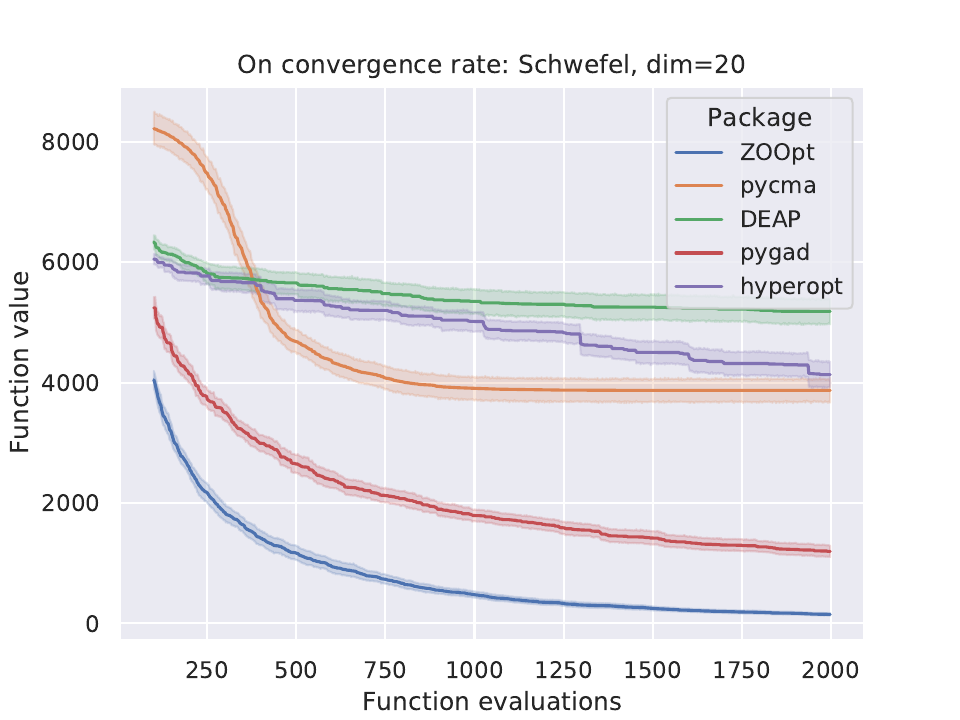}
	\end{minipage}
\begin{minipage}[t]{\picspace \linewidth}
	\centering
	\includegraphics[width=\expwidth in]{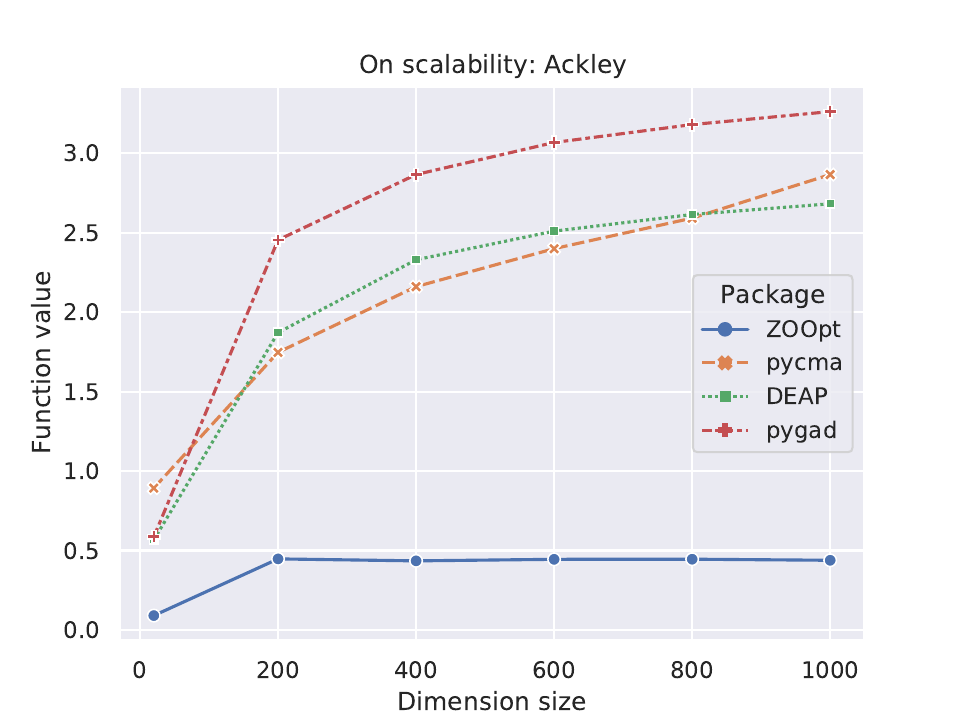}
\end{minipage}%
\begin{minipage}[t]{\picspace \linewidth}
	\centering
	\includegraphics[width=\expwidth in]{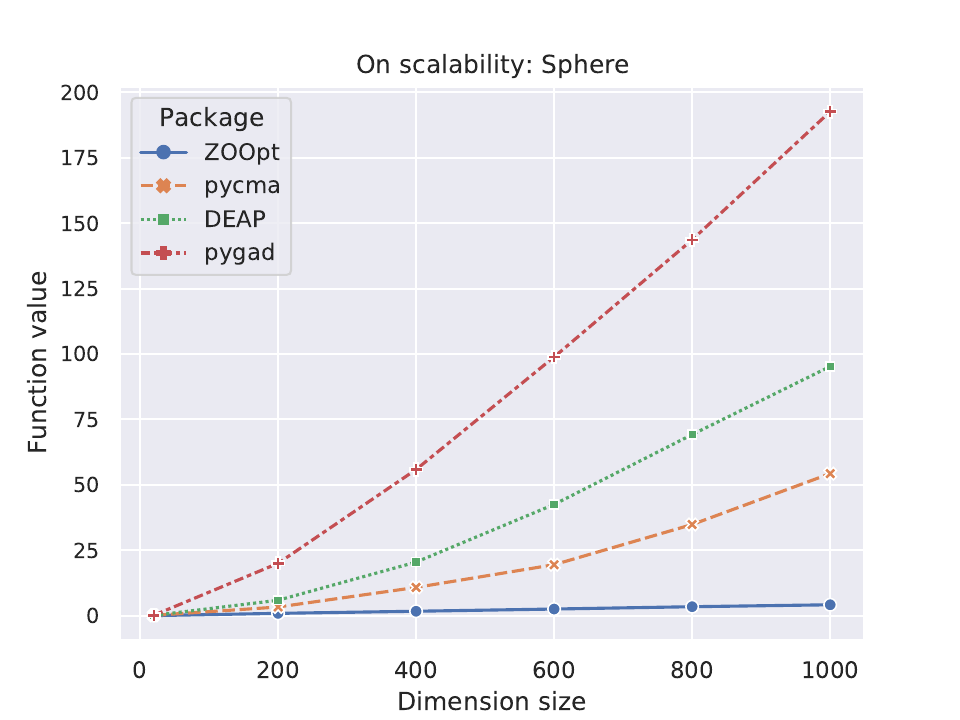}
\end{minipage}
\begin{minipage}[t]{\picspace\linewidth}
	\centering
	\includegraphics[width=\expwidth in]{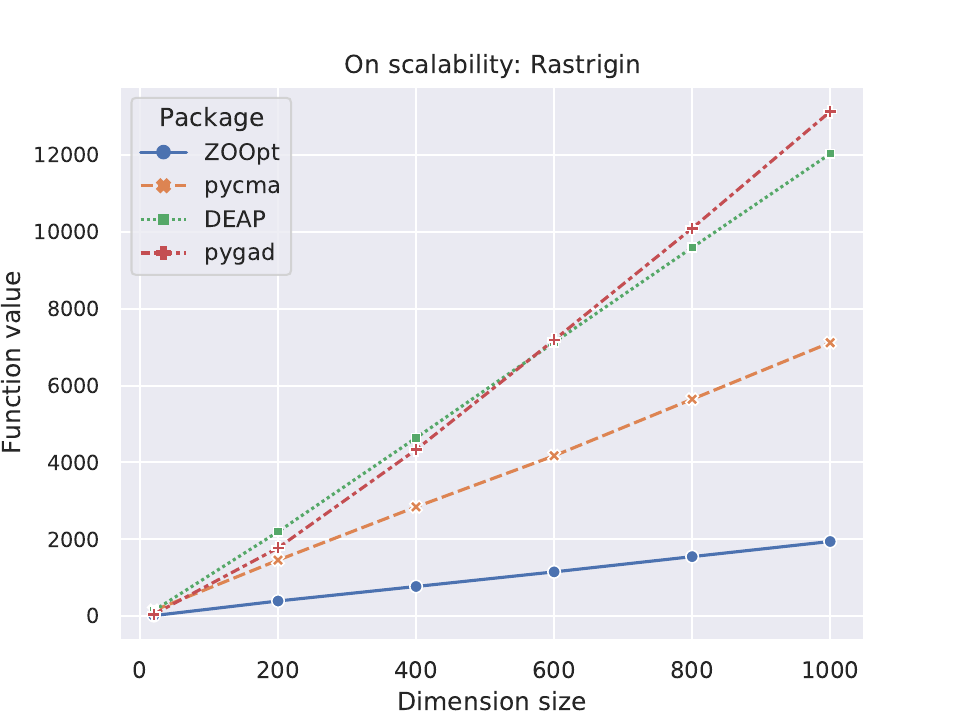}
\end{minipage}
\begin{minipage}[t]{\picspace \linewidth}
	\centering
	\includegraphics[width=\expwidth in]{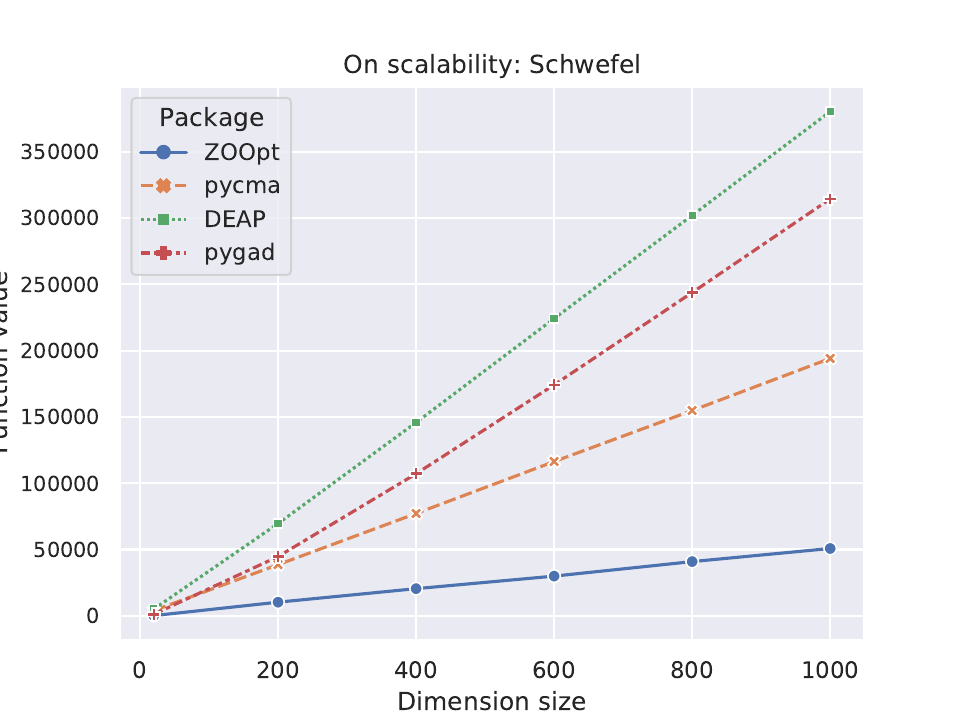}
\end{minipage}
\begin{minipage}[t]{\picspace \linewidth}
	\centering
	\includegraphics[width=\expwidth in]{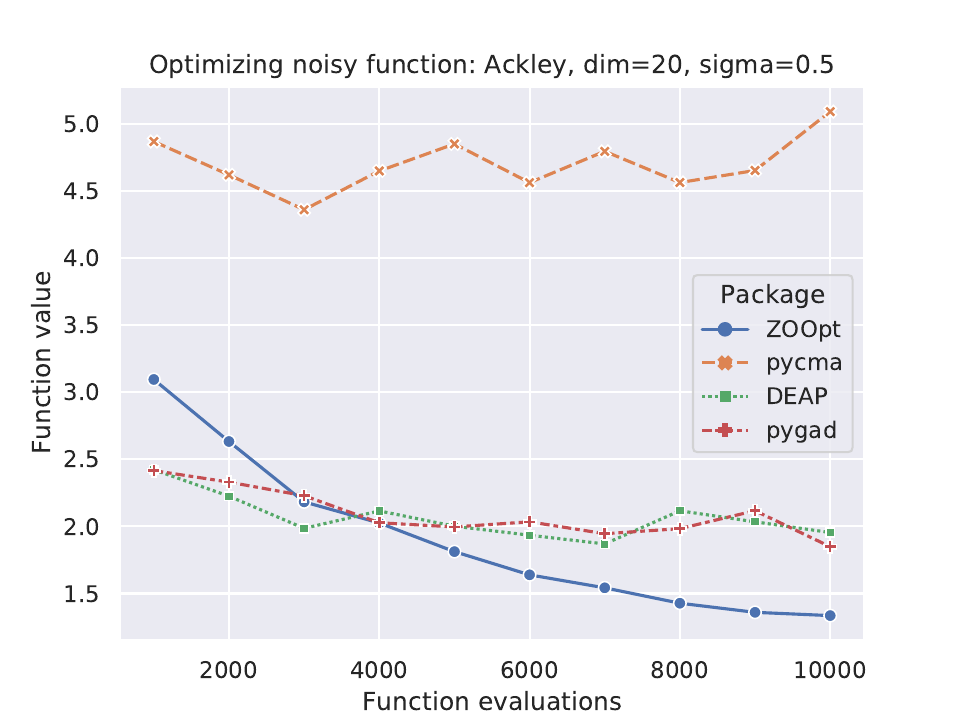}
\end{minipage}%
\begin{minipage}[t]{\picspace \linewidth}
	\centering
	\includegraphics[width=\expwidth in]{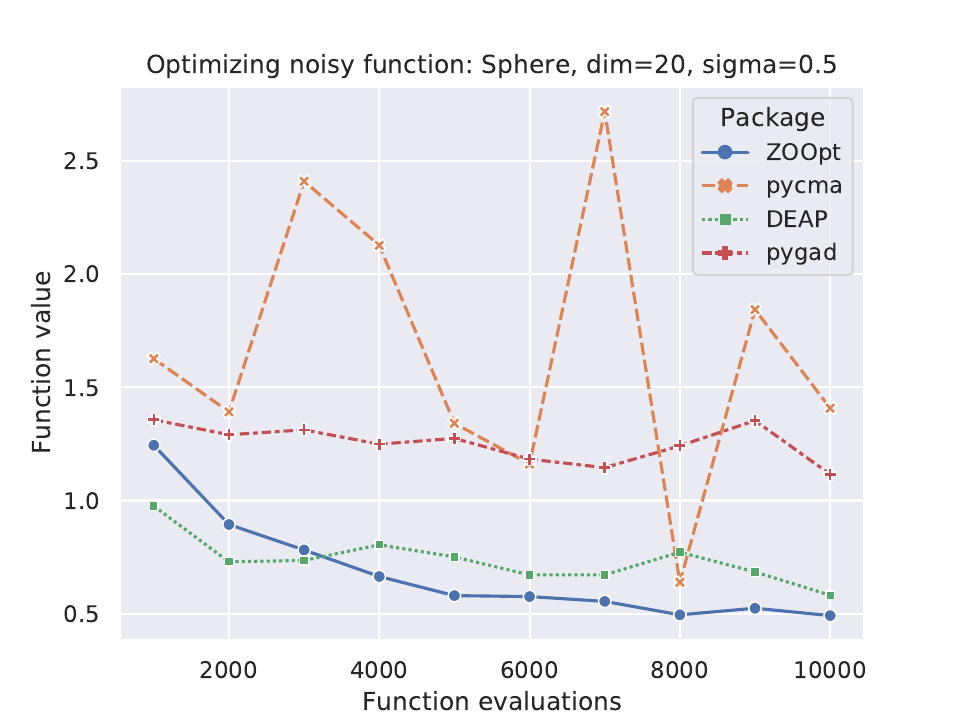}
\end{minipage}
\begin{minipage}[t]{\picspace\linewidth}
	\centering
	\includegraphics[width=\expwidth in]{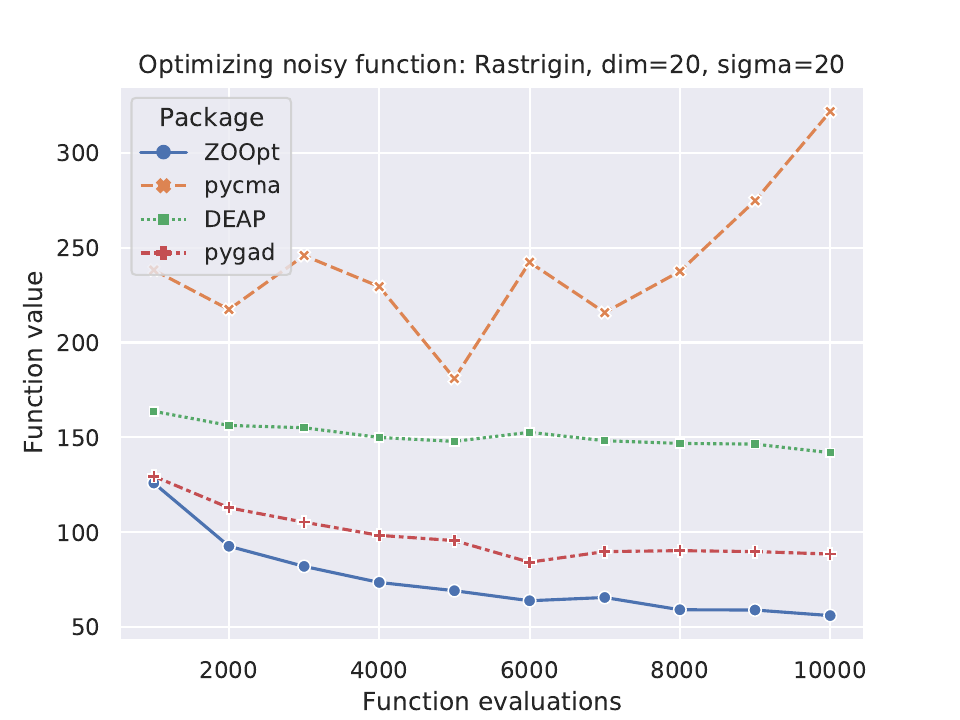}
\end{minipage}
\begin{minipage}[t]{\picspace \linewidth}
	\centering
	\includegraphics[width=\expwidth in]{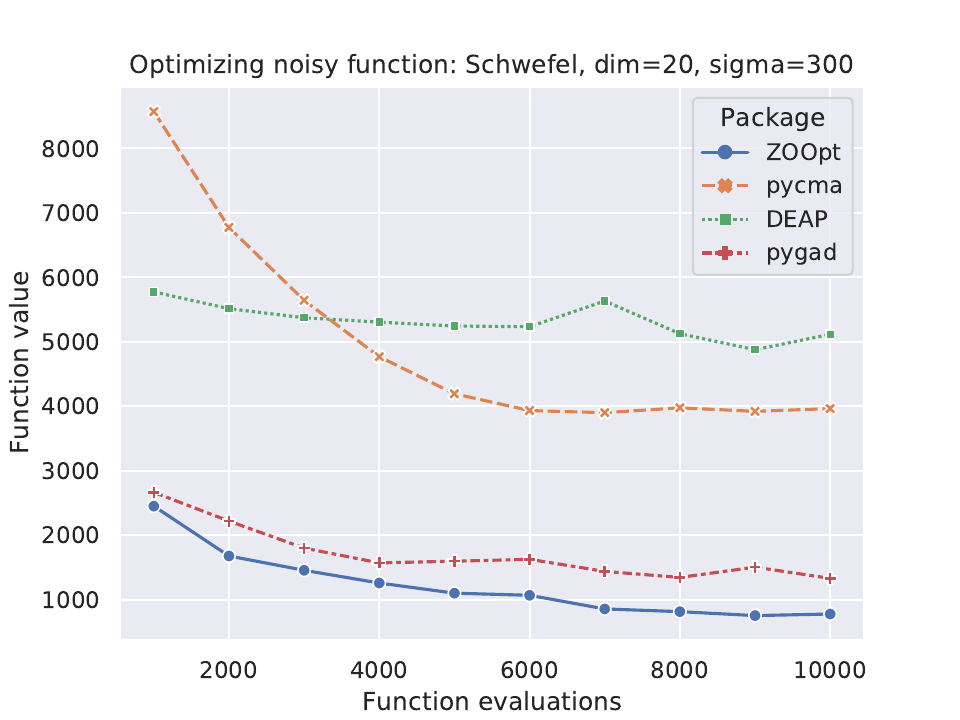}
\end{minipage}

	\caption{Evaluate the optimization (minimization) performance of ZOOpt on four benchmark synthetic functions. The top row shows  the convergence rate of the tested toolboxes. The middle row shows the scalibility of the tested toolboxs  as the dimension size increases. The bottom row demonstrates the performance on optimizing noisy functions. It can be observed that ZOOpt achieves the best performance in all tasks.}
	\label{figure1}
\end{figure*}

To answer question (1), (2), (3), we conduct experiments on optimizing benchmark synthetic functions. Among them, the Ackley, Rastrigin and Schwefel functions are highly non-convex while the Sphere function is convex. The optimal values of four functions are all zero.  The Ackley and Sphere functions are minimized within the seach space $X =  [-1, 1]^d$, where $d$ is the dimension size. The Rastrigin function is minimied within $[-5, 5]^d$. The Schwefel function is minimized within $[-500, 500]^d$.  The optimal position of each function (except the Schwefel function, which is fixed to $[420.97, ..., 420.97]$) is shifted from $[0, ..., 0]$ to a random point sampled from $[0.2\ *\ l, 0.2\ *\ u]^d$, where $l$ and $u$ respectively refer to the lower and upper bound of the search space on that dimension. This is to avoid a possible optimization bias to the origin point. The 3-d  graphs of these functions are shown in Figure \ref{sfunctions}. Each experiment is repeated for 30 times. Mean values and 95\%  confidence intervals are recorded. Results are shown in Figure \ref{figure1}.

\textbf{On convergence rate.} We set the dimension size to be 20 for each objective function and number of evaluations to be 2000. We study the convergence rate with regard to the number of function evaluations by recording the best-so-far solution value during the optimization. As shown in the top row of Figure \ref{figure1}, ZOOpt reduces the objective function value with the highest rate in all tasks.

\textbf{On Scalability} The scalability of derivative-free optimization methods is critical on solving large-scale problems. In this experiment, we quantitatively study the scalability of ZOOpt. We set the dimension size $d$ to be 20, 200, 400, 600, 800, 1000 and the number of function evaluations to be $100 \times d$. The confidence interval is omitted for clarity. The middle row of Figure \ref{figure1} shows that ZOOpt has the lowest growth rate on the function value in all tasks as the dimension size increases, indicating that ZOOpt has better scalability than other toolboxes.

\textbf{On robustness against noise.} To study  the performance of ZOOpt on optimizing noisy object, we add the Gaussian noise to original functions to simulate the noisy environment. The new objective functions are defined as $f^N (x) = f (x) + N (0, \sigma^2)$.  The number of function evaluations is set to be  10000. For all tasks, ZOOpt and pycma use their built-in noise handler while DEAP and pygad not. It can be observed that ZOOpt reduces the function value at a steady pace as the number of evaluation increases despite the noise.

\subsection{Results on classification tasks with Ramploss.}
The Ramp loss is defined as
$R_s(z) = H_1(z) - H_s(z)$ with $s < 1$, where $H_s(z) = \max\{0, s - z\}$ is the Hinge loss with $s$ being the Hinge point.
The task is to find a vector $w$ and a scalar $b$ to minimize  $f(w, b)=\frac{1}{2}\|w\|_{2}^{2}+C \sum_{\ell}^{L} R_{s}\left(y_{\ell}\left(w^{\top} v_{\ell}+b\right)\right)$, where $v_l$ is the training instance and $y_l \in \{-1, +1\}$ is its label. Due to the convexity of the Hinge loss, the number of support vectors increases linearly with
the number of training instances in SVM, which is undesired with respect to scalability. While this problem can be relieved by using the Ramp loss ~\citep{DBLP:conf/icml/CollobertSWB06}. 

We employ two binary class UCI datasets, Adult and Bank, for the classification task. Discrete variables of the original features are preprocessed by one-hot encoding. Continuous variables are normalized into $[-1, 1]$. The result feature dimension (excluding the label) is expanded to 108 for Adult and 51 for Bank. Since we focus on the optimization performance, we only compare the results on the complete data set. Two hyper-parameters, i.e. C and s, are adjustable in the optimization formulation. We set $s \in \{-1, 0\}$ and $C \in \{0.1, 0.5, 1, 2, 5, 10\}$  to study the effectiveness of the tested toolboxes under different hyper-parameters. We set the total number of calls to the objective function to be 40n for all toolboxes. The achieved objective values are reported in Table \ref{ramploss_table}.

It can be observed that ZOOpt is comparable with pycma and dominate DEAP and pygad in all cases. Notice that the smaller the $C$ is, the closer the objective function is to convexity. Therefore, the optimization difficulty increases with $C$. Although the results of ZOOpt and pycma are close, ZOOpt achieves the better results when $C$ is large, i.e., the objective function is further from the convexity. Pycma is better when the objective function is closer to the convexity. 

\begin{table}[t]
	\begin{minipage}[t]{\linewidth}
		\centering
		\includegraphics[width=\textwidth ]{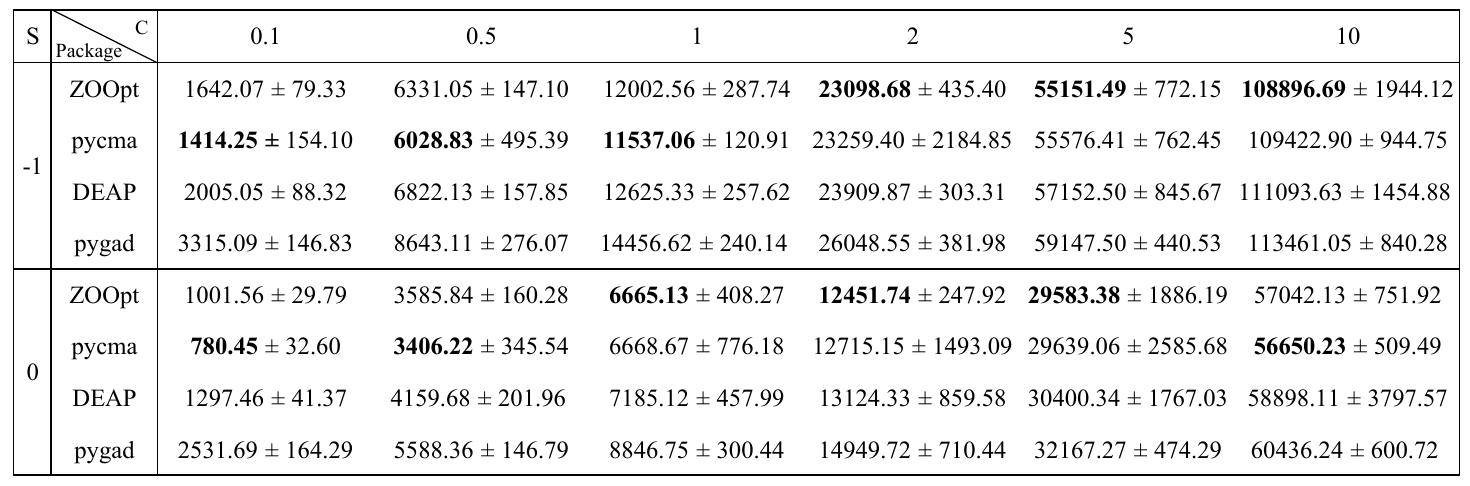}
	\end{minipage}
	\begin{minipage}[t]{\linewidth}
		\centering
		\includegraphics[width=\textwidth]{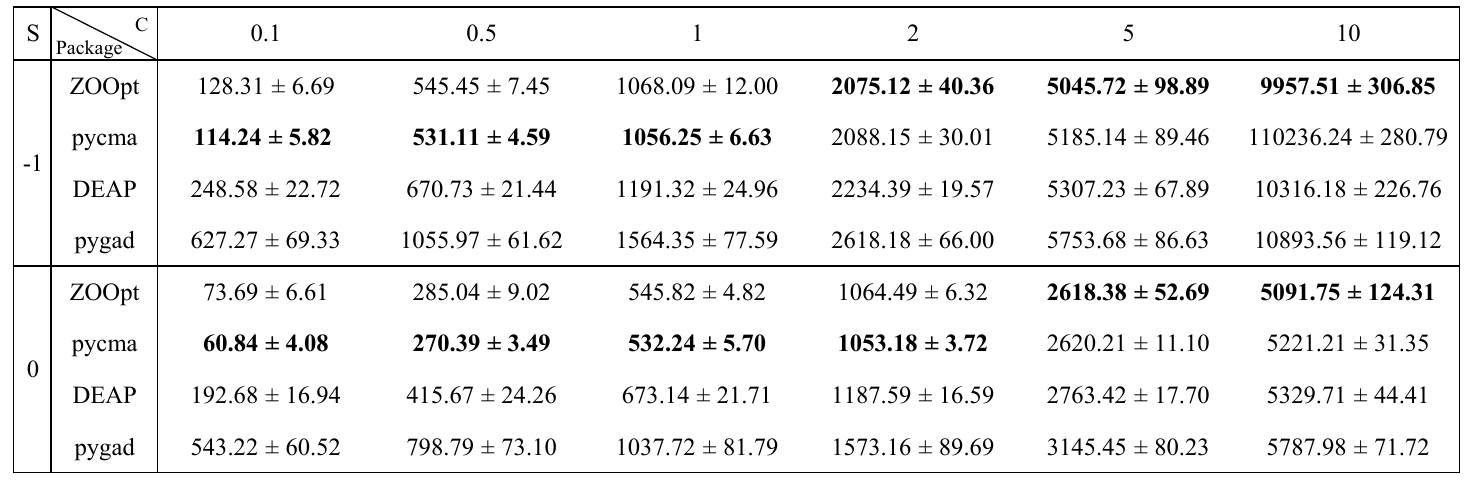}
	\end{minipage}
	\caption{Results on the Adult (upper) and Bank (lower) data sets. Comparing the achieved objective function values against the parameter C of the classification with Ramp loss.}
    \label{ramploss_table}
\end{table}

\subsection{Results on direct policy search for OpenAI controlling tasks.}
\textbf{Gym tasks.} In the OpenAI Gym environment, we use
six existing controlling tasks, ‘Acrobot’, ‘MountainCar’,
‘HalfCheetah’, ‘Hopper’, 'Humanoid' and 'Swimmer', to test the toolboxes. We apply the feedforward
neural network as the policy. The task information and neural network structures are showed in Table \ref{dps_task_description}. For example, in ‘Acrobot’: $|S| = 6$, $|A| = 3$; the neural network has two hidden layers with 5 and 3 neurons each; $|w| = 48$; the activation function for hidden layers and the output layer are respectively relu and softmax; the maximum number of steps is 500. We will give a summary of each task. More details can be found in the homepage of OpenAI Gym. In ‘Acrobot’, system includes two joints and two links, where the joint between the two links is actuated. Initially, the links are hanging downwards and the goal of this task is to swing the end of the low link up to a given height. In ‘MountainCar’, a car is positioned in a valley between two mountains and wants to drive up the mountain on the right by building up momentum. 'HalfCheetah', 'Hopper', 'Humanoid' and ‘Swimmer'  are simulation tasks. In those tasks, policy control simulated objects to achieve a goal. For example, in ‘HalfCheetah’, policy should
control a cheetah with half body running forward as fast as possible. The tasks of 'Acrobot' and 'MountainCar' are finding policies with smallest step number when goals are met. The tasks except for ‘Acrobot’ and ‘MountainCar’ are finding policies to control object getting scores from the environment as high as possible. Therefore, in Table \ref{dps}, columns of 'Acrobot' and 'MountainCar' are step numbers, the smaller the better. The other rows are the cumulative rewards from environments, the larger the better. 
\begin{table}[t]
	\centering
	\includegraphics[width=0.95\textwidth ]{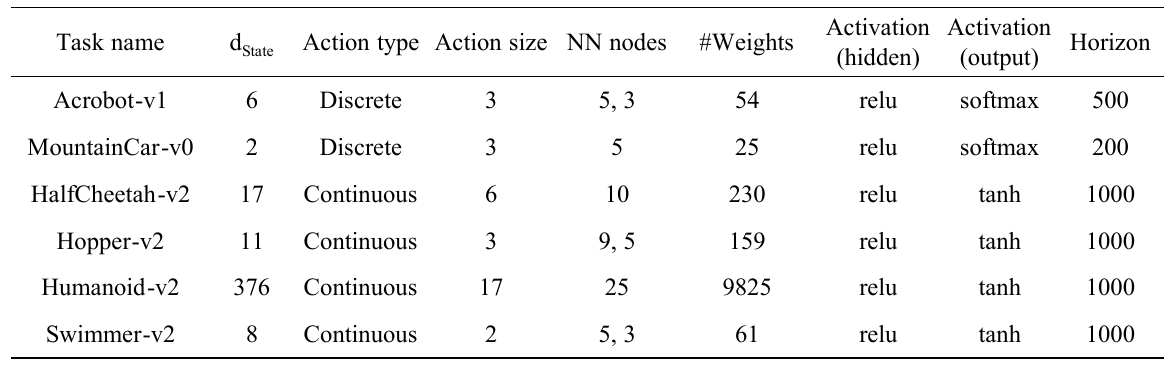}
	\caption{The parameters of the direct policy search for OpenAI controlling tasks.}
    \label{dps_task_description}
\end{table}

The average cumulative rewards of 10 simulations is used as the evaluation value of a neural network to reduce noise. The solution space $X$ is set to be $[-10, 10]^{\text{\#Weight}}$. The output of the neural network is scaled to be within the action space, which is defined by the environment. All toolboxes use 2,000 evaluations for each task. The best solution will be re-evaluated for 30 times to reduce the noise further and their mean value will be recorded as the final result. Each experiment is repeated for 10 times. The mean value and the standard deviation are recorded in Table \ref{dps}. It can be observed that ZOOpt obtained the best results on 5/6 tasks.
\begin{table}[htb]
	\begin{minipage}[t]{\linewidth}
		\centering
		\includegraphics[width=\textwidth ]{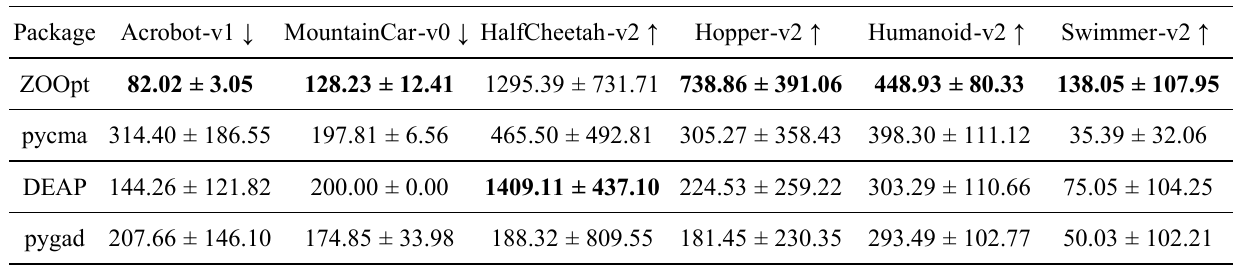}
	\end{minipage}
	\caption{ The mean scores and the standard deviation of the best found policy by each toolbox. The numbers in bold mean the best scores in each column. The mark ↓ means the score is the smaller the better, and ↑ means the larger the better.}
	\label{dps}
\end{table}

\section{Conclusion}
In this paper, we introduce the toolbox ZOOpt which provides efficient derivative-free solvers and is designed easy to use. By combining several state-of-the-art classification-based optimization methods, noise-handlers and high-dimensionality handlers, ZOOpt is particularly good at optimization problems in machine learning. By incorporating with Ray, the optimization in ZOOpt can be easily distributed across multiple machines. In empirical studies, we firstly study the convergence rate, the scalability and the robustness against noise of ZOOpt on optimizing synthetic functions. ZOOpt achieves the best performance in all of these experiments. We then test ZOOpt on two machine learning tasks. Results on classification tasks with Ramploss show that ZOOpt is comparable with pycma and dominates other toolboxes. Results on  direct policy search for OpenAI controlling tasks show that ZOOpt achieved the best performance on 5/6 tasks. For a detailed tutorial of the usage of ZOOpt, we refer readers to the project homepage \url{https://github.com/polixir/ZOOpt}.

\acks{We would like to acknowledge the support for this project from the National Key Laboratory for Novel Software Technology at Nanjing University (No. KFKT2021B14), the Natural Science Foundation of Shanghai (No. 21ZR1420300), Shanghai Key Laboratory of Multidimensional Information Processing at East China Normal University (No. MIP202101), and the Fundamental Research Funds for the Central Universities. }

\vskip 0.2in
\bibliographystyle{unsrt}
\bibliography{sample}

\begin{thebibliography}{24}
\providecommand{\natexlab}[1]{#1}
\providecommand{\url}[1]{\texttt{#1}}
\expandafter\ifx\csname urlstyle\endcsname\relax
  \providecommand{\doi}[1]{doi: #1}\else
  \providecommand{\doi}{doi: \begingroup \urlstyle{rm}\Url}\fi

\bibitem[Bergstra et~al.(2013)Bergstra, Yamins, and
  Cox]{DBLP:conf/icml/BergstraYC13}
J.~Bergstra, D.~Yamins, and D.~D. Cox.
\newblock Making a science of model search: Hyperparameter optimization in
  hundreds of dimensions for vision architectures.
\newblock In \emph{Proceedings of the 30th International Conference on Machine
  Learning}, volume~28, pages 115--123, Atlanta, GA, 2013.

\bibitem[Collobert et~al.(2006)Collobert, Sinz, Weston, and
  Bottou]{DBLP:conf/icml/CollobertSWB06}
R.~Collobert, F.~H. Sinz, J.~Weston, and L.~Bottou.
\newblock Trading convexity for scalability.
\newblock In \emph{Proceedings of the 23rd International Conference on Machine
  Learning}, volume 148, pages 201--208, Pittsburgh, Pennsylvania, 2006.

\bibitem[de~Boer et~al.(2005)de~Boer, Kroese, Mannor, and Rubinstein]{Boer2005}
P.~de~Boer, D.~P. Kroese, S.~Mannor, and R.~Y. Rubinstein.
\newblock A tutorial on the cross-entropy method.
\newblock \emph{Annals of Operations Research}, 134\penalty0 (1):\penalty0
  19--67, 2005.

\bibitem[Feurer et~al.(2015)Feurer, Klein, Eggensperger, Springenberg, Blum,
  and Hutter]{NIPS15autosklearn}
M.~Feurer, A.~Klein, K.~Eggensperger, J.~T. Springenberg, M.~Blum, and
  F.~Hutter.
\newblock Efficient and robust automated machine learning.
\newblock In \emph{Advances in Neural Information Processing Systems 28}, pages
  2962--2970, Montreal, Canada, 2015.

\bibitem[Fortin et~al.(2012)Fortin, {De Rainville}, Gardner, Parizeau, and
  Gagn\'e]{DEAP_JMLR2012}
F.-A. Fortin, F.-M. {De Rainville}, M.-A. Gardner, M.~Parizeau, and C.~Gagn\'e.
\newblock {DEAP}: Evolutionary algorithms made easy.
\newblock \emph{Journal of Machine Learning Research}, 13:\penalty0 2171--2175,
  2012.

\bibitem[Gad(2021)]{gad2021pygad}
A.~F. Gad.
\newblock Pygad: An intuitive genetic algorithm python library.
\newblock \emph{CoRR}, abs/2106.06158, 2021.

\bibitem[Hansen et~al.(2003)Hansen, M{\"{u}}ller, and Koumoutsakos]{HansenMK03}
N.~Hansen, S.~D. M{\"{u}}ller, and P.~Koumoutsakos.
\newblock Reducing the time complexity of the derandomized evolution strategy
  with covariance matrix adaptation {(CMA-ES)}.
\newblock \emph{Evolutionary Computation}, 11\penalty0 (1):\penalty0 1--18,
  2003.

\bibitem[Hansen et~al.(2019)Hansen, Akimoto, and Baudis]{hansen2019pycma}
N.~Hansen, Y.~Akimoto, and P.~Baudis.
\newblock {CMA-ES/pycma} on {G}ithub.
\newblock Zenodo, DOI:10.5281/zenodo.2559634, 2019.

\bibitem[Hu et~al.(2017)Hu, Qian, and Yu]{hu2017sequential}
Y.-Q. Hu, H.~Qian, and Y.~Yu.
\newblock Sequential classification-based optimization for direct policy
  search.
\newblock In \emph{Proceedings of the 31st AAAI Conference on Artificial
  Intelligence}, pages 2029--2035, San Francisco, CA, 2017.

\bibitem[Liu et~al.(2019)Liu, Hu, Qian, and Yu]{Asracos2019}
Y.~Liu, Y.~Hu, H.~Qian, and Y.~Yu.
\newblock Asynchronous classification-based optimization.
\newblock In \emph{Proceedings of the First International Conference on
  Distributed Artificial Intelligence}, pages 9:1--9:8, Beijing, China, 2019.

\bibitem[Moritz et~al.(2018)Moritz, Nishihara, Wang, Tumanov, Liaw, Liang,
  Elibol, Yang, Paul, Jordan, and Stoica]{MoritzNWTLLEYPJ18}
P.~Moritz, R.~Nishihara, S.~Wang, A.~Tumanov, R.~Liaw, E.~Liang, M.~Elibol,
  Z.~Yang, W.~Paul, M.~I. Jordan, and I.~Stoica.
\newblock Ray: {A} distributed framework for emerging {AI} applications.
\newblock In \emph{13th {USENIX} Symposium on Operating Systems Design and
  Implementation}, pages 561--577, Carlsbad, CA, 2018.

\bibitem[Munos(2014)]{MunosFTML2014}
R.~Munos.
\newblock From bandits to {Monte-Carlo Tree Search}: The optimistic principle
  applied to optimization and planning.
\newblock \emph{Foundations and Trends in Machine Learning}, 7\penalty0
  (1):\penalty0 1--130, 2014.

\bibitem[Qian et~al.(2015)Qian, Yu, and Zhou]{qian.yu.nips15}
C.~Qian, Y.~Yu, and Z.-H. Zhou.
\newblock Subset selection by pareto optimization.
\newblock In \emph{Advances in Neural Information Processing Systems 28}, pages
  1765--1773, Montreal, Canada, 2015.

\bibitem[Qian et~al.(2016{\natexlab{a}})Qian, Shi, Yu, Tang, and
  Zhou]{DBLP:conf/ijcai/QianSYTZ16}
C.~Qian, J.~Shi, Y.~Yu, K.~Tang, and Z.~Zhou.
\newblock Parallel pareto optimization for subset selection.
\newblock In S.~Kambhampati, editor, \emph{Proceedings of the 25th
  International Joint Conference on Artificial Intelligence}, pages 1939--1945,
  New York, NY, 2016{\natexlab{a}}. {IJCAI/AAAI} Press.

\bibitem[Qian et~al.(2017)Qian, Shi, Yu, Tang, and Zhou]{NIPS17noise}
C.~Qian, J.-C. Shi, Y.~Yu, K.~Tang, and Z.-H. Zhou.
\newblock Subset selection under noise.
\newblock In \emph{Advances in Neural Information Processing Systems 30}, pages
  3563--3573, Long Beach, CA, 2017.

\bibitem[Qian et~al.(2016{\natexlab{b}})Qian, Hu, and Yu]{IJCAI16sre}
H.~Qian, Y.-Q. Hu, and Y.~Yu.
\newblock Derivative-free optimization of high-dimensional non-convex functions
  by sequential random embeddings.
\newblock In \emph{Proceedings of the 25th International Joint Conference on
  Artificial Intelligence}, pages 1946--1952, New York, NY, 2016{\natexlab{b}}.

\bibitem[Real et~al.(2017)Real, Moore, Selle, Saxena, Suematsu, Tan, Le, and
  Kurakin]{ICicml17}
E.~Real, S.~Moore, A.~Selle, S.~Saxena, Y.~L. Suematsu, J.~Tan, Q.~V. Le, and
  A.~Kurakin.
\newblock Large-scale evolution of image classifiers.
\newblock In \emph{Proceedings of the 34th International Conference on Machine
  Learning}, pages 2902--2911, Sydney, Australia, 2017.

\bibitem[Salimans et~al.(2017)Salimans, Ho, Chen, and Sutskever]{SalimansHCS17}
T.~Salimans, J.~Ho, X.~Chen, and I.~Sutskever.
\newblock Evolution strategies as a scalable alternative to reinforcement
  learning.
\newblock \emph{CoRR}, abs/1703.03864, 2017.

\bibitem[Shahriari et~al.(2016)Shahriari, Swersky, Wang, Adams, and
  de~Freitas]{reviewBO16}
B.~Shahriari, K.~Swersky, Z.~Wang, R.~P. Adams, and N.~de~Freitas.
\newblock Taking the human out of the loop: {A} review of {B}ayesian
  optimization.
\newblock \emph{Proceedings of the {IEEE}}, 104\penalty0 (1):\penalty0
  148--175, 2016.

\bibitem[Snoek et~al.(2012)Snoek, Larochelle, and Adams]{SnoekLA2012}
J.~Snoek, H.~Larochelle, and R.~P. Adams.
\newblock Practical {B}ayesian optimization of machine learning algorithms.
\newblock In \emph{Advances in Neural Information Processing Systems 25}, pages
  2960--2968, Lake Tahoe, NV, 2012.

\bibitem[Thornton et~al.(2013)Thornton, Hutter, Hoos, and
  Leyton{-}Brown]{KDD13autoweka}
C.~Thornton, F.~Hutter, H.~H. Hoos, and K.~Leyton{-}Brown.
\newblock {A}uto-{WEKA}: {C}ombined selection and hyperparameter optimization
  of classification algorithms.
\newblock In \emph{Proceedings of the 19th {ACM} {SIGKDD} International
  Conference on Knowledge Discovery and Data Mining}, pages 847--855, Chicago,
  IL, 2013.

\bibitem[Wang et~al.(2018)Wang, Qian, and Yu]{AAAI18noise}
H.~Wang, H.~Qian, and Y.~Yu.
\newblock Noisy derivative-free optimization with value suppression.
\newblock In \emph{Proceedings of the 32nd AAAI Conference on Artificial
  Intelligence}, New Orleans, LA, 2018.

\bibitem[Wang et~al.(2016)Wang, Zoghi, Hutter, Matheson, and
  Freitas]{Wang2016remboJAIR}
Z.~Wang, M.~Zoghi, F.~Hutter, D.~Matheson, and N.~D. Freitas.
\newblock Bayesian optimization in a billion dimensions via random embeddings.
\newblock \emph{Journal of Artificial Intelligence Research}, 55:\penalty0
  361--387, 2016.

\bibitem[Yu et~al.(2016)Yu, Qian, and Hu]{yu.qian.racos}
Y.~Yu, H.~Qian, and Y.-Q. Hu.
\newblock Derivative-free optimization via classification.
\newblock In \emph{Proceedings of the 30th AAAI Conference on Artificial
  Intelligence}, pages 2286--2292, Phoenix, AZ, 2016.

\end{thebibliography}

\end{document}